\documentclass[letterpaper, 10 pt, journal, twoside]{IEEEtran}

\ifCLASSINFOpdf
  \usepackage[pdftex]{graphicx}

\else

\fi

\usepackage{amsmath}
\usepackage{amsfonts}
\usepackage{booktabs}
\usepackage{tabularx}
\usepackage{multirow}
\usepackage[table]{xcolor}
\usepackage{colortbl}
\usepackage{makecell}
\usepackage{booktabs}
\usepackage{multirow} 
\usepackage{cite}
\newcommand{\SE}{$SE(3)$ }
\newcommand{\stopgrad}[1]{\mathrm{sg}[#1]}

\newcommand{\rebuttle}[1]{#1}

\begin{document}

\title{Vision-Based End-to-End Learning for UAV Traversal of Irregular Gaps via Differentiable Simulation}

\author{Linzuo Zhang$^*$\thanks{$^*$ These authors contributed equally to this work.}, Yu Hu$^*$, Feng Yu, Yang Deng, Wenxian Yu, Danping Zou$^{\dag}$\thanks{$^{\dag}$ Corresponding author.}
\thanks{The authors are with the Shanghai Jiao Tong University, Shanghai 200240, China.}
\thanks{Digital Object Identifier (DOI): see top of this page.}
\thanks{\scriptsize \tt Emails: \{zhanglinzuo, henryhuyu, yu-feng, dengyang24, dpzou\}@sjtu.edu.cn}
}


\maketitle

\begin{abstract}
Navigation through narrow and irregular gaps is an essential skill in autonomous drones for applications such as inspection, search-and-rescue, and disaster response. However, traditional planning and control methods rely on explicit gap extraction and measurement, while recent end-to-end approaches often assume regularly shaped gaps, leading to poor generalization and limited practicality. In this work, we present a fully vision-based, end-to-end framework that maps depth images directly to control commands, enabling drones to traverse complex gaps within unseen environments. Operating in the \textit{Special Euclidean group} \SE, where position and orientation are tightly coupled, the framework leverages differentiable simulation, a Stop-Gradient operator, and a Bimodal Initialization Distribution to achieve stable traversal through consecutive gaps. Two auxiliary prediction modules—a gap-crossing success classifier and a traversability predictor—further enhance continuous navigation and safety. Extensive simulation and real-world experiments demonstrate the approach’s effectiveness, generalization capability, and practical robustness.
\end{abstract}

\section{Introduction}

Unmanned aerial vehicles (UAVs) are agile agents capable of dynamic flight in three-dimensional environments. In real-world missions, from disaster relief to industrial inspection, they must often navigate cluttered spaces and pass through irregular openings, requiring rapid reconfiguration of body orientation. 
However, most vision-based navigation methods~\cite{loquercio_learning_2021, de2020collision, redder2025vds, yu2024mavrl, Zhang2024BackTN, Hu2024seeingpixelmotionlearning, lee2025quadrotor} neglect position--attitude coupling, rather than considering in the full \SE state space. This limitation becomes critical for traversing narrow and tilted gaps that demand precise orientation control.

Traditional SE(3) planning and control methods~\cite{2017Aggressive,2018SearchBased} rely on constrained optimization with feedback control and require accurate prior knowledge of the environment. 
Learning-based end-to-end policies for \SE gap traversal aim to map sensory inputs directly to control commands. These reinforcement and imitation learning approaches show promise but have inherent limitations. Specifically, most methods~\cite{Xiao2023FlyingThru, Chen2023Responsive, Lin2019Flyingthrough} require known gap poses or geometric cues to extract low-dimensional states, while certain image-to-control approaches~\cite{2021WholeBody} rely on image masks for feature extraction. Such methods struggle to traverse irregular gaps common in real-world environments—non-rectangular, uneven, or partially occluded—because the absence of clear corners and edges hinders feature extractors from inferring optimal traversal poses. 
Fully vision-based end-to-end methods for narrow-gap traversal remain under-explored despite their strong potential for robust real-world deployment.
\begin{figure}[t]
    \centering
    \includegraphics[width=\linewidth]{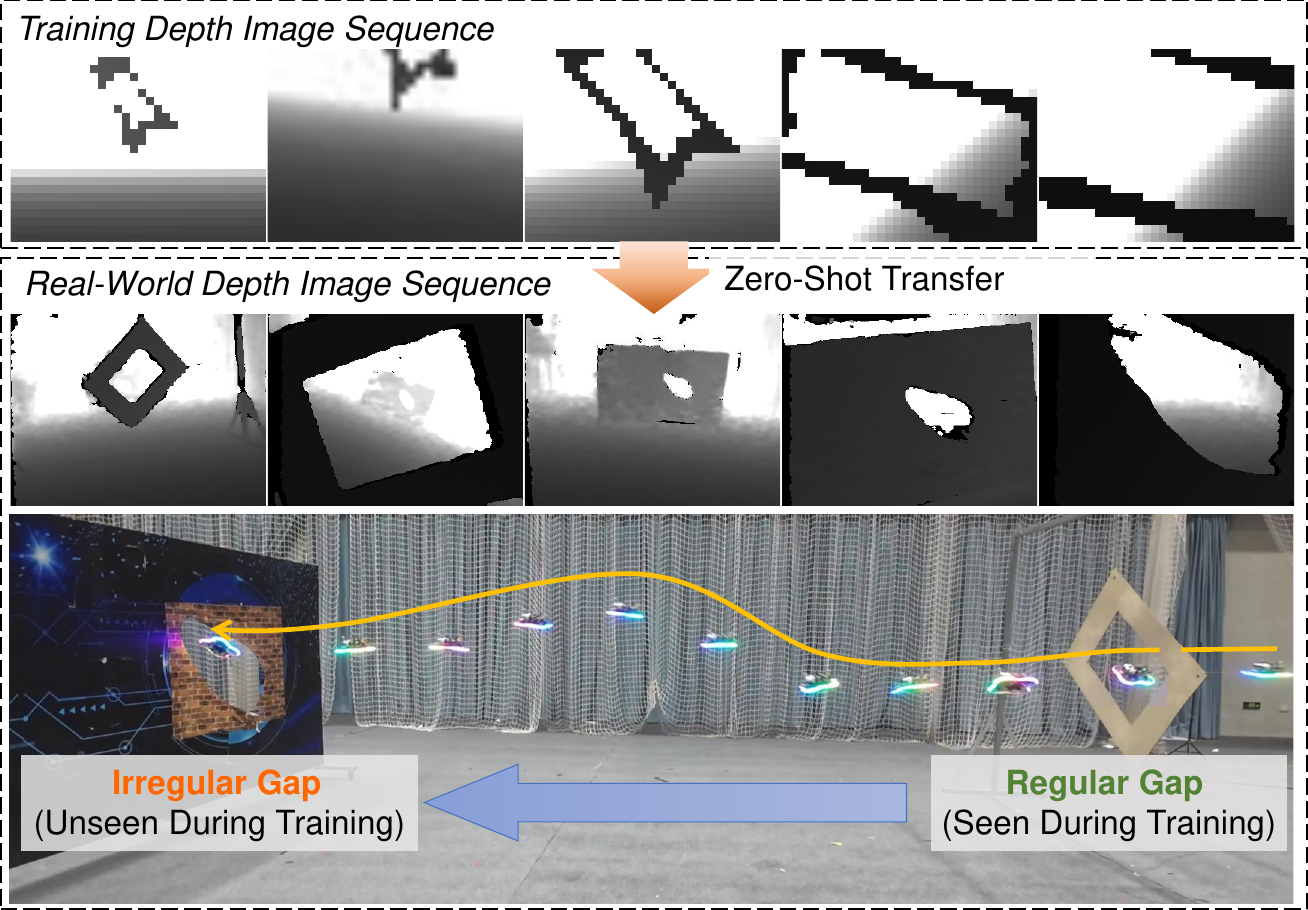} 
    \caption{{\bf Visualization of the training and real-world evaluation:} The top row shows depth image sequences collected during training; the middle row shows depth images captured by the real drone during execution; the bottom row illustrates the real-world trajectory, where the drone first flies through a regular gate similar to the training scenarios and then successfully navigates an irregular, previously unseen gap, demonstrating the policy’s generalization ability.}
    \label{fig:fig1}
    \vspace{-1.5em}
\end{figure}

Differentiable simulation has emerged as a promising approach for learning vision-based, end-to-end policies as it provides precise and informative gradient signals. In this context, differentiable simulation refers to a framework where system dynamics are differentiable, allowing task losses to be back-propagated through time for gradient-based policy optimization. This paradigm enables efficient learning through system dynamics and improves generalization from simple training environments to complex real-world scenarios~\cite{Zhang2024BackTN, Hu2024seeingpixelmotionlearning}. While effective in obstacle avoidance, its use for tasks requiring high-precision \SE control remains largely unexplored.


\begin{figure*}[th] 
  \centering
  \includegraphics[width=\textwidth] {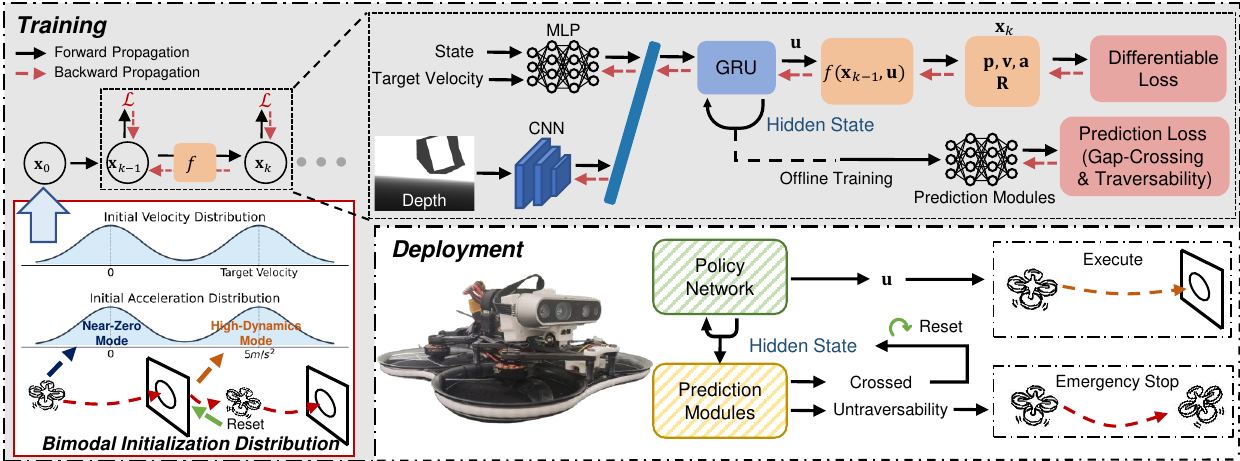} 
    \caption{\textbf{System Overview.} An end-to-end policy maps depth images directly to control commands and is trained via differentiable simulation, enabling direct back-propagation of task losses to the network. A gap-crossing detection module resets the policy hidden state to support continuous multi-gap traversal, while a traversability prediction module improves safety in challenging environments. A bimodal initialization distribution stabilizes training and enhances robustness across successive gaps.}

   \vspace{-1.0em}
  \label{fig:system}
\end{figure*} 

In this work, we leverage differentiable simulation to address \SE planning and control tasks. We propose a fully vision-based, end-to-end framework that maps depth images directly to control commands, enabling drones to traverse narrow and irregularly shaped gaps including configurations not encountered during training (Fig.~\ref{fig:fig1}). To this end, we employ a Stop-Gradient (SG) operator to selectively block certain gradients and adopt a Bimodal Initialization Distribution to stabilize consecutive gap traversal and post-gap attitude control. To further enhance practical applicability, we introduce two complementary data-driven prediction modules: a gap-crossing success classifier that facilitates continuous multi-gap navigation with stable flight, and a traversability predictor that assesses traversal feasibility based on the policy’s hidden state, allowing the system to anticipate potential hazards.

In summary, our main contributions are:

\begin{enumerate}
\item A practical and deployable fully vision-based, end-to-end \SE policy that generalizes from regular training gaps to irregular real-world openings, inherently robust to perception noise and outperforming traditional RL approaches that rely on explicit perception preprocessing.
\item The design of auxiliary prediction modules for gap-crossing detection and traversability assessment, which enhance reliability in continuous navigation.
\item Extensive validation in both simulation and real-world experiments, demonstrating the feasibility, robustness, and generalization capability of the proposed approach.
\end{enumerate}

\section{Related Work} 

\subsection{Flight Through Gaps} 
Trajectory planning and tracking pipelines are widely used for SE(3) quadrotor control. Falanga et al.~\cite{2017Aggressive} studied gap traversal via ballistic trajectories, while~\cite{yang2021whole} formulated constrained trajectory optimization using a point-mass model within a convex hull. Liu et al.~\cite{2018SearchBased} proposed a map-based coarse-to-fine planner with hierarchical refinement. Although these methods leverage differential flatness to simplify planning, they rely on prior knowledge and are sensitive to perception noise, limiting real-world robustness.

\rebuttle{
Recent vision-based end-to-end navigation methods map depth images directly to control commands, demonstrating strong performance in obstacle avoidance and goal-directed flight~\cite{loquercio_learning_2021, de2020collision, redder2025vds, yu2024mavrl}. However, extending these approaches to agile \SE gap traversal remains challenging.}
Existing learning-based gap traversal methods~\cite{Xiao2023FlyingThru, Chen2023Responsive, Lin2019Flyingthrough} require additional modules to extract target information from images, such as known gap poses or geometric cues—which restricts generalization. Wu et al.~\cite{2021WholeBody} adopt a reinforcement–imitation learning framework with direct image inputs; however, their approach relies on auxiliary image masks, which introduce additional prior assumptions and require dedicated preprocessing, leading to a more complex pipeline and limiting scalability to real-world applications.

In contrast, our method directly maps raw depth images to control commands, enabling fully vision-based, robust multi-gap traversal with strong generalization to unseen environments.

\subsection{Differentiable Simulation}

Differentiable simulation has emerged as a powerful tool for training end-to-end policies through system dynamics~\cite{metz_gradients_2021,Wiedemann2023APG}, showing strong results in real-world quadrotor~\cite{Heeg2024hover,li2025visfly} and quadruped~\cite{Song2024learning,wiedemannwueest2023training} applications. While effective due to low-variance first-order gradients, these studies often underexplore visual inputs.

Recent studies~\cite{Zhang2024BackTN,Hu2024seeingpixelmotionlearning} integrate CUDA-based rendering for training, achieving impressive sim-to-real transfer. However, their simulators are restricted to simple primitives (e.g., spheres, cubes), limiting scalability to complex structures such as random gaps. Differentiable simulators~\cite{li2025visfly,zhang2025diffaero,kulkarni2025} can generate richer objects at high speed, but object shapes and appearances remain predefined, constraining flexibility and randomization.

Moreover, the dynamics in~\cite{Zhang2024BackTN,Hu2024seeingpixelmotionlearning} are simplified as point-mass systems, sufficient for agile flight but untested for tasks requiring large-angle attitude control or precise position–attitude coupling. A recent extension~\cite{lee2025quadrotor} introduces a Collective Thrust and Body Rate (CTBR) formulation, yet still focuses on simple point-to-point navigation without demanding tight position–attitude interaction.

These limitations motivate our approach. We combine differentiable simulation with a mesh-based depth renderer to flexibly generate complex gap geometries, and employ a CTBR dynamics model for precise \SE control. Together, these enable fully vision-based, end-to-end gap traversal in previously unseen environments.

\section{Problem Statement}\label{STATEMENT}

Fig.~\ref{fig:system} shows the overall framework, which maps raw depth images to low-level control commands for \SE gap traversal without explicit perception or planning. A gap-crossing prediction module resets policy hidden states after each traversal, while a traversability predictor enhances safety. During training, a bimodal initialization exposes the policy to both stable and high-velocity post-traversal states, enabling rapid stabilization across successive gaps.

\subsection{Problem Formulation}
We consider learning a control policy $\pi_\theta$ for a quadrotor operating in the full six-degree-of-freedom (6-DoF) space of rigid-body motions, formally described by the Special Euclidean group \(\mathrm{SE}(3)\). The system state at time $t$ is
\begin{align}
    \mathbf{x}_{t} = \left(\mathbf{p}_t, \mathbf{R}_t, \mathbf{v}_t, \mathbf{a}_t, \boldsymbol{\omega}_t\right),
\end{align}
where $\mathbf{p}_t \in \mathbb{R}^3$ is the position, $\mathbf{R}_t \in \mathrm{SO}(3)$ is the rotation matrix, $\mathbf{v}_t \in \mathbb{R}^3$ is the linear velocity, $\mathbf{a}_t \in \mathbb{R}^3$ is the linear acceleration, and $\boldsymbol{\omega}_t \in \mathbb{R}^3$ is the angular velocity. The system evolves according to
\begin{align}
    \mathbf{x}_{t+1} = f(\mathbf{x}_t, \mathbf{u}_t),
\end{align}
where $\mathbf{u}_t$ denotes the control input.

At each time step, the policy network takes three inputs: a raw depth image $o_t$, a partial quadrotor state $\mathbf{s}_t$ (body-frame velocity and the second column of the rotation matrix), and a target velocity vector $\mathbf{\Tilde{v}}_\text{target}$ directed toward the gap with the desired speed. To avoid treating this vector as a strict reference, large random perturbations (±2 m) are added so that it serves only as a guidance signal. The policy then produces control commands
\begin{align}
\mathbf{u}_t = \pi_\theta(o_t, \mathbf{s}_t, \mathbf{\Tilde{v}}_\text{target}),
\end{align}
enabling reactive and stable gap traversal.

The objective is to find the optimal policy parameters $\theta^\ast$ by minimizing the cumulative control loss over a horizon of length $N$:
\begin{align}
    \mathcal{L}_{\theta} &= \sum_{t=0}^{N-1} l(\mathbf{x}_t, \pi_{\theta}(o_t, \mathbf{s}_t^\text{b}, \mathbf{\Tilde{v}}_\text{target})), \\
    \theta &\leftarrow \theta - \gamma \nabla_{\theta} \mathcal{L}_\theta,
\end{align}
where $\gamma$ is the learning rate. Since both the dynamics $f$ and the loss $l$ are differentiable, gradients can be back-propagated through the simulated system into the visual encoder, enabling end-to-end training of perception and control.

\subsection{Quadrotor Dynamics}
We define the control input at time $t$ as 
$\mathbf{u}_t = \left(\boldsymbol{\omega}_{c,t}, c_{c,t}\right)$, 
where $\boldsymbol{\omega}_{c,t}$ and $c_{c,t}$ denote the commanded body rates and collective thrust, respectively. 
We adopt a differentiable Collective Thrust and Body Rate (CTBR) model to propagate the quadrotor state in discrete time. 
Commanded body rates and thrust are modeled with a first-order response:
\begin{align}
    \boldsymbol{\omega}_t &= \alpha_{\omega}\boldsymbol{\omega}_{t-1} + (1-\alpha_{\omega})\boldsymbol{\omega}_{c,t}, \\
    c_t &= \alpha_{c}c_{t-1}+ (1-\alpha_{c})c_{c,t},
\end{align}
where $\alpha_{\omega} = \exp(-\Delta t/\tau_{\omega})$ and $\alpha_{c} = \exp(-\Delta t/\tau_{c})$ are exponential decay factors determined by the time constants $\tau_{\omega}$ and $\tau_{c}$, respectively. 
The rotation is updated using the exponential map:
\begin{align}
    \mathbf{R}_{t+1} &= \mathbf{R}_t \exp\left(\left[\boldsymbol{\omega}_t\right]_{\times} \Delta t\right),
\end{align}
and the acceleration is computed as
\begin{align}
    \mathbf{a}_t &= \frac{1}{m}\left(c_t \mathbf{R}_t \mathbf{e}_3 - m g \mathbf{e}_3\right) - k_v \mathbf{v}_t.
\end{align}
Finally, velocity and position are updated via Euler integration:
\begin{align}
    \mathbf{v}_{t+1} &= \mathbf{v}_t + \mathbf{a}_t \Delta t, \\
    \mathbf{p}_{t+1} &= \mathbf{p}_t + \mathbf{v}_t \Delta t + \frac{1}{2}\mathbf{a}_t \Delta t^2.
\end{align}
Here, $m$ is the mass of the quadrotor, $g$ is the gravitational acceleration, $k_v$ is a linear drag coefficient, and $\mathbf{e}_3 = [0, 0, 1]^\top$.

The parameters $\tau_{\omega}$, $\tau_{c}$, and $k_v$ are identified from real flight data, and are randomly perturbed by a factor of $0.9\!\sim\!1.1$ during training to improve robustness.

\section{Policy Optimization via Differentiable Simulation}\label{sec:diffsim}
To optimize the control policy, we construct a differentiable simulation environment that integrates a CTBR quadrotor dynamics model with a mesh-based depth renderer (Fig.~\ref{fig:renderer}). Depth images rendered from the environment serve as policy observations, while gradients from the task loss flow backward through the dynamics to update the policy parameters. This end-to-end differentiability allows joint optimization of visual perception and control.

\subsection{Mesh-based Depth Rendering} 
\begin{figure}[ht]
  \centering
  \includegraphics[width=\linewidth]{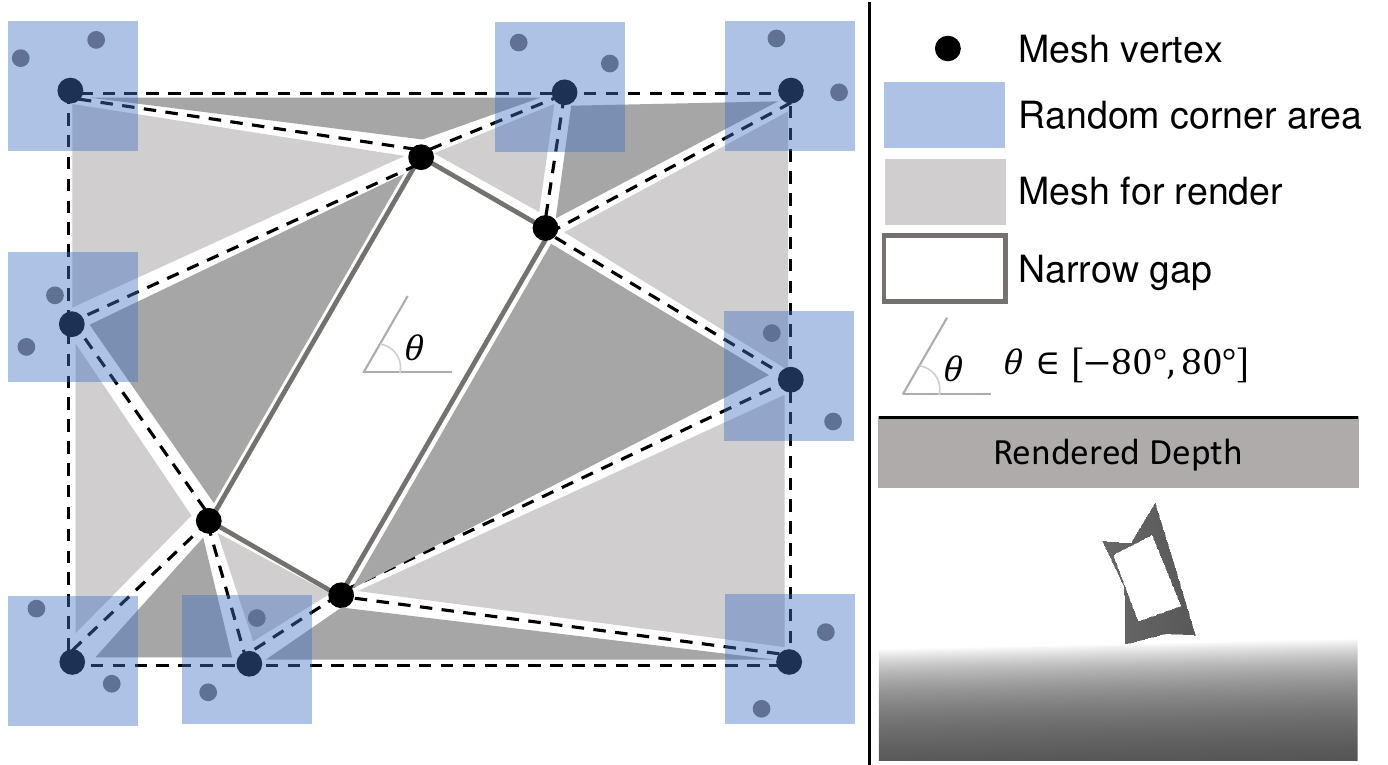}
  \caption{\textbf{Mesh-based depth renderer.} A high-speed CUDA-based renderer generates depth images from mesh geometries. Domain randomization is applied by perturbing the angles and corner vertices, producing diverse gap configurations for training.}
  \label{fig:renderer}
  \vspace{-1.0em}
\end{figure}

We use a CUDA-accelerated renderer to generate depth images from 3D mesh environments composed of triangular faces. Narrow gaps are modeled with twelve vertices(Fig.~\ref{fig:renderer}). Depth values are obtained by casting rays from the camera origin and computing their intersections with mesh surfaces. To enhance generalization, domain randomization is applied by perturbing frame and midpoint vertices within bounded regions and uniformly sampling gap orientations in the range \(-80^{\circ}\) to \(80^{\circ}\). This approach creates diverse outer aperture geometries without altering the underlying mesh structure, enabling efficient parallelized rendering for high-throughput simulation.

Although training is conducted using standard rectangular gaps for efficiency, the learned policy represents a continuous mapping in the high-dimensional visual input space, rather than relying on explicit geometric templates. This property enables natural generalization to irregularly shaped openings at deployment.

\subsection{Back-propagation through Time}

\rebuttle{In contrast to other CUDA-accelerated simulators such as Isaac Sim\footnote{https://github.com/isaac-sim/IsaacSim}, which solely support forward simulation and do not provide analytical gradients,} 
our simulator constructs a computational graph with \emph{automatic differentiation}, enabling policy optimization via Backpropagation Through Time (BPTT).
Trajectories are generated by iteratively applying policy commands, after which gradients propagate from the task loss through the dynamics to the policy parameters.

\begin{equation}
  \small
       \frac{1}{N}\sum_{k=0}^{N-1} \left( \sum_{i=0}^{k} \frac{\partial l_k }{\partial x_{k}} \prod_{j=i+1}^{k} \left( \frac{\partial x_{j}}{\partial x_{j-1}} e^{-\alpha \Delta t} \right) \frac{\partial x_i}{\partial \theta} + \frac{ \partial l_k}{\partial u_k} \frac{ \partial u_k}{\partial \theta} \right),
\end{equation}
where $\alpha$ is an exponential decay factor applied to dynamics gradients. 
\subsection{Loss Function}  
We define the gap traversal task to ensure safe and efficient passage while maintaining smooth and stable flight. The loss function consists of six terms: position error \(\mathcal{L}_p\), rotation error \(\mathcal{L}_r\), forward alignment \(\mathcal{L}_f\), forward velocity tracking \(\mathcal{L}_v\), action smoothness \(\mathcal{L}_a\), and jerk smoothness \(\mathcal{L}_j\).  

The traversal position loss is defined as  
\begin{equation}
\label{pos_loss}
\mathcal{L}_p = \big\| \mathbf{p}_k^{y,z \mid \text{g}} - \mathbf{p}_\text{g}^{y,z \mid \text{g}} \big\| 
\cdot  \stopgrad{\max\left(1 - d_k^{\text{g}}, 0\right)},
\end{equation}  
where $\mathbf{p}_k^{y,z \mid \text{g}}$ and $\mathbf{p}_\text{g}^{y,z \mid \text{g}}$ are the quadrotor's and gap's $y,z$-coordinates projected onto the plane of the gap, respectively, and $d_k^{\text{g}}$ is the distance from the quadrotor to the gap plane. Here, $\stopgrad{\cdot}$ denotes the Stop-Gradient (SG) operator, which prevents gradients from propagating through the distance term. This ensures that the position loss focuses on critical moments near traversal while avoiding spurious gradients.

Inspired by~\cite{Lee2010geometric}, we define the rotation loss as:
\begin{equation}
\label{rot_loss}
\mathcal{L}_r = 
\Big\| 
\frac{1}{2} 
\left( 
\mathbf{R}_\text{g}^\top \mathbf{R}_k - \mathbf{R}_k^\top \mathbf{R}_\text{g}
\right)^\vee 
\Big\|
\cdot 
\stopgrad{\max\left(1 - d_k^{\text{g}}, 0\right)},
\end{equation}
where \(\mathbf{R}_\text{g}\) and \(\mathbf{R}_k\) denote the rotation matrices of the gap and the quadrotor, respectively, and \((\cdot)^\vee\) denotes the mapping of a skew-symmetric matrix to a vector.  

The  operator is applied to the distance term to block gradient flow, preventing spurious updates—particularly at large tilt angles—that could drive the policy toward local minima, cause hesitation, or produce discontinuous motion, as shown in our ablation study. Applying SG to both position and rotation losses stabilizes the learning process and enables smooth, accurate gap traversal.

Forward velocity tracking is 
\begin{equation} 
    \mathcal{L}_v =  \| \mathbf{v}_k - \mathbf{v}_{\text{ref}} \| \cdot \stopgrad{b_k}, 
\end{equation} 
where $\mathbf{v}_{\text{ref}}$ is the reference speed along the gap normal and $b_k=1$ if the quadrotor has not passed the gap plane, 0 otherwise. 

Forward alignment to the gap center is 
\begin{equation} 
    \mathcal{L}_f = \arccos\left( \frac{\mathbf{p}_\text{g} - \mathbf{p}_k}{\|\mathbf{p}_\text{g} - \mathbf{p}_k\|} \cdot \mathbf{e}_1 \right) \cdot \stopgrad{b_k}. 
\end{equation} 
which penalizes deviations of the drone's forward axis (x-axis) from the direction of the gap center. This encourages the quadrotor to always face the gap during traversal, ensuring that the onboard camera maintains a clear view of the target. 

Action and jerk smoothness losses are 
\begin{equation} 
    \mathcal{L}_a = \frac{1}{T}\sum_{k=1}^T \|\mathbf{u}_k\|^2, \quad \mathcal{L}_j = \frac{1}{T-1}\sum_{k=1}^{T-1} \Big\| \frac{\mathbf{u}_k - \mathbf{u}_{k+1}}{\Delta t} \Big\|^2. 
\end{equation}

The total loss is the weighted sum
\begin{equation}
\mathcal{L} = \lambda_p \mathcal{L}_p + \lambda_r \mathcal{L}_r + \lambda_v \mathcal{L}_v 
+ \lambda_f \mathcal{L}_f + \lambda_a \mathcal{L}_a + \lambda_j \mathcal{L}_j,
\label{equ:loss}
\end{equation}
with empirically chosen weights \(\lambda_p=10\), \(\lambda_r=10\), \(\lambda_v=0.1\), \(\lambda_f=1.0\), \(\lambda_a=0.01\), and \(\lambda_j=0.0001\).

\subsection{Bimodal Initialization Distribution}
In real deployments, large accelerations caused by aggressive gap crossings make post-traversal control highly challenging. Directly constraining post-gap states during training would introduce gradients at the traversal moment in BPTT, potentially leading to overly conservative behavior. At the same time, we aim to avoid increasing environment complexity, so that a policy trained in a single-gap environment can be directly transferred to multi-gap navigation.

To achieve this, during training we initialize the quadrotor’s state with a bimodal distribution, illustrated in Fig.~\ref{fig:system}, with one peak near zero for stable first-gap traversal and another near the target velocity with large acceleration,  encouraging the policy to rapidly stabilize under challenging post-traversal dynamics. During deployment, in combination with our Gap-Crossing Prediction Module, the hidden state is reset after each successful traversal, effectively treating the post-gap state as a fresh initial condition. This allows the policy to remain stable after each traversal, thereby enabling seamless transfer from single-gap training to robust multi-gap navigation.

\section{Auxiliary Prediction Modules}

The goal of this work is to build a policy-driven framework. We adopt a data-driven strategy while incorporating prior knowledge, enabling the network to explicitly perceive both successful crossings and traversal feasibility. To this end, we introduce two complementary prediction modules: a gap-crossing classifier, which detects traversal events, and a traversability predictor, which estimates the feasibility of passing through narrow gaps. Both operate solely on the policy network’s recurrent hidden state, which compactly encodes task-relevant dynamics and past states, avoiding redundant visual processing while retaining strong predictive capability.

\paragraph*{\textbf{Gap-Crossing Classifier Module}}
This module monitors progress with respect to each gap and detects successful traversals. Upon detection, it resets the policy’s recurrent hidden state to its initial value. Implemented as a lightweight MLP, it takes the hidden state as input and outputs a binary indicator of traversal success. Trained offline with simulation trajectories and optimized via binary cross-entropy, it enables continuous multi-gap navigation with stable behavior.

\paragraph*{\textbf{Traversability Prediction Module}}
This module estimates whether the quadrotor can safely pass narrow gaps by producing a scalar traversability score from the hidden state. Labels are generated in simulation by scaling gap sizes to balance positive and negative samples. At each step, trajectories are labeled traversable if all minimum distances between the quadrotor and the gap exceed a safety threshold. Also trained offline with binary cross-entropy, the module anticipates safe traversal from orientation, position, velocity and gap's size providing feedback for corrective actions and improving real-world reliability.
\section{Experiments}
We evaluate the proposed end-to-end quadrotor control framework through extensive experiments in both simulation and real-world environments. Our evaluation covers four aspects: ablation studies in simulation to quantify the contributions of key design choices, real-world tests to validate sim-to-real transfer and assess zero-shot generalization to previously unseen environments, verification of the traversability prediction module to assess its reliability, and analysis of policy robustness under ambiguous target velocity inputs. Collectively, these experiments demonstrate the effectiveness, safety, and strong generalization capabilities of the proposed approach.

\subsection{Architecture and Training Details}
\rebuttle{
\subsubsection{Network Architecture}
The control policy is structured as a vision-based recurrent network. Specifically, a single-channel depth image ($32\times24$) is first transformed into inverse depth and then downsampled using a $2\times2$ max-pooling operation. The resulting feature map is processed by a 3-layer convolutional neural network (CNN) (channels: $[32, 64, 128]$, kernels: $[2, 3, 3]$, strides: $[2, 1, 1]$) with LeakyReLU activations to yield a $192$-D visual embedding. This embedding is fused with a projected $192$-D state vector via element-wise addition. The fused features are then integrated by a single-layer gated recurrent unit (GRU), which feeds into a linear head for continuous control. Additionally, the GRU hidden state serves as input to auxiliary multilayer perceptron (MLP) predictors, each consisting of a single hidden layer (hidden size: $1024$).
}

\subsubsection{Training Performance}
We evaluate the end-to-end policy and auxiliary prediction modules in our CUDA-based differentiable simulator. Training runs on an Intel i9 CPU with an Nvidia RTX 4090 GPU, using AdamW ($\text{lr}=1\times10^{-3}$, batch size 64) for 50,000 iterations. Each iteration includes 80-step rollouts, with target velocities uniformly sampled from 2--4~m/s. Gaps are procedurally generated 3--5~m ahead, with lateral offsets from $[-1,1]$~m and heights from $[1,2]$~m, yielding diverse geometries. Training finishes in \textbf{44 minutes}. With the highly parallelized mesh-based depth renderer, the simulator renders depth maps at \textbf{1.45M frames/s}.

Prediction modules are trained in the same simulator but with reduced minimum gap sizes, decreased from 0.8~m $\times$ 0.4~m to 0.5~m $\times$ 0.25~m. During this stage, pretrained policy weights are fixed and gradients are blocked through the dynamics. Using the same optimizer and hyperparameters, training runs for 50,000 iterations with 70-step rollouts, completing in \textbf{19 minutes}.

\begin{figure}[t]
    \centering
    \includegraphics[width=\linewidth]{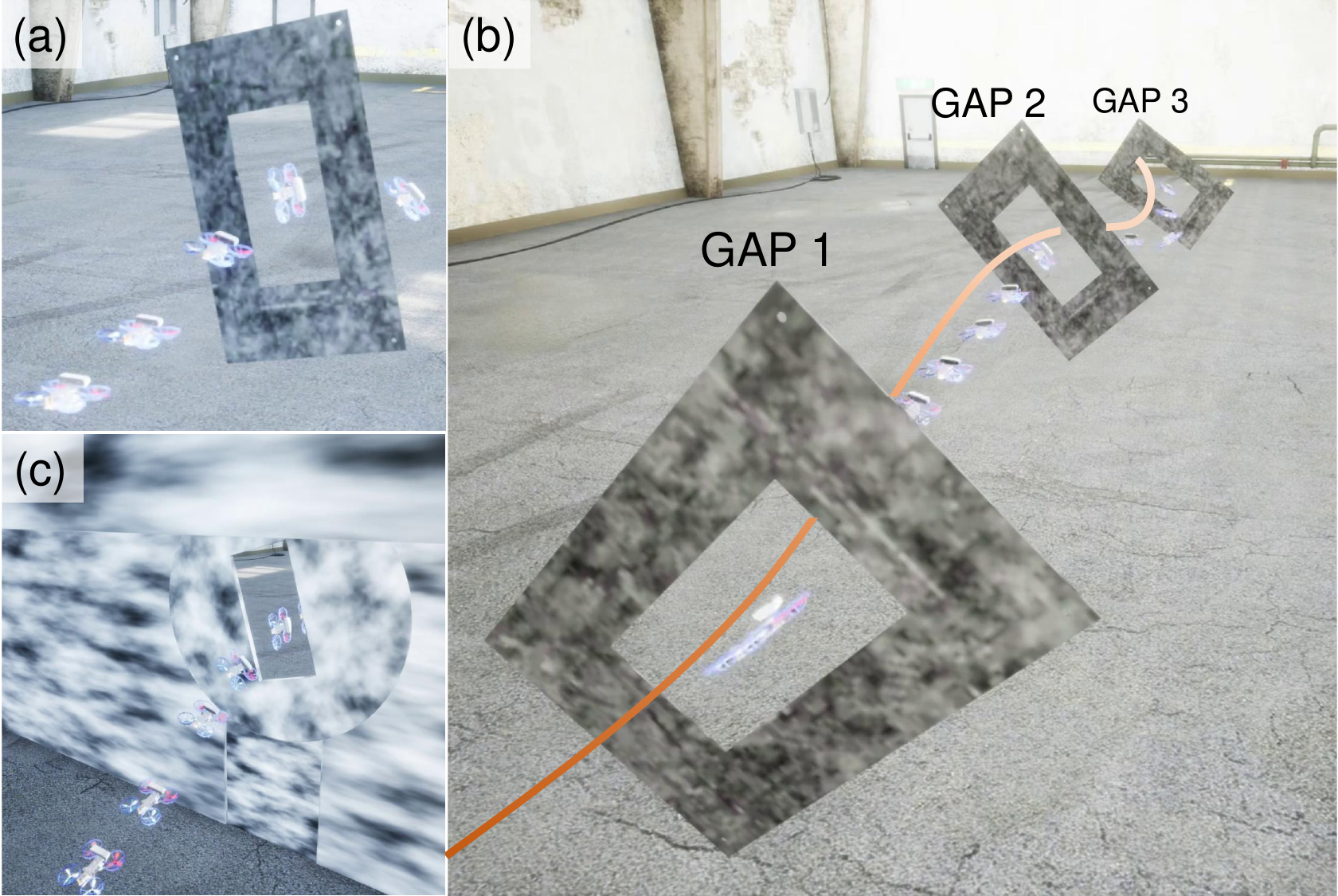}
    \caption{\rebuttle{\textbf{AirSim Simulation environments.} (a) Single-gap scenario, where the quadrotor flies through a single tilted gap. (b) Multi-gap scenario, where the quadrotor sequentially flies through multiple tilted gaps. (c) Wall-mounted gap scenario, where a square opening is embedded in a planar wall.}}
    \label{fig:simulation}
    \vspace{-0.8em}
\end{figure}
\subsection{Sim-to-Sim Transfer}
We conducted a series of experiments using the \rebuttle{AirSim simulator \cite{airsim2017fsr}} to quantify the contributions of the proposed design components. Our testing scenarios included:
\begin{enumerate}
    \item \textbf{Single-gap scenario}: the gap (0.9~m in length and 0.4~m in width) was placed 3--5~m in front of the quadrotor, with lateral offsets sampled from $[-0.5, 0.5]$~m, heights from $[1.5, 1.7]$~m, and tilt angles from $-80^\circ$ to $80^\circ$, as shown in Fig.~\ref{fig:simulation} (a).
    \rebuttle{In addition, we consider a wall-mounted gap scenario, where a square opening is embedded in a planar wall, as illustrated in Fig.~\ref{fig:simulation}(c).}
    \item \textbf{Multi-gap scenario}: each gap was spaced 3--5~m apart, with lateral offsets from $[-1, 1]$~m and tilt angles from $-50^\circ$ to $50^\circ$, as shown in Fig.~\ref{fig:simulation} (b).
\end{enumerate}

To better approximate real-world conditions, we captured left and right grayscale images and applied the Semi-Global Matching(SGM) algorithm~\cite{Hirschmuller2007} to obtain depth maps with more realistic noise characteristics. In addition, because the exact positions of the gates were unknown, the network input was defined as $\mathbf{\Tilde{v}}_{\text{target}} = v_{d} \mathbf{e}_{1}$ where $v_{d}$ is an externally specified desired velocity magnitude.

\subsubsection{Baseline Comparison}
In AirSim, similar to~\cite{2023LearningAgileOnboard,Xiao2023FlyingThru}, we trained a state-based policy using RL, specifically the PPO~\cite{schulman2017proximal} algorithm, which directly takes the target position and attitude as input. A front-end perception module based on an Edge Drawing algorithm \cite{Topal2011edge} was designed to extract the target position and orientation from depth images. Traversal success is defined as passing through the gap without any collisions.

\rebuttle{
During training, the PPO agent maximizes a reward defined as the negative of the loss function used in Eq.(\ref{equ:loss}), i.e., $r=-\mathcal{L}$. The policy network consists of a GRU-based encoder followed by an MLP. The agent is trained under the same environment and setting as our method.
}

Based on this criterion, we first compared the success rates of the two systems using real depth images, where both our method and the RL baseline achieved strong performance (98\% vs. 80\%). However, when the input images were replaced with SGM depth maps, which severely degraded the performance of the front-end Edge Drawing algorithm, the RL baseline exhibited a substantial drop in success rate due to its sensitivity to noise in the perception module, whereas our method experienced only a minor decrease. These results demonstrate that our end-to-end policy is inherently robust to perception noise, outperforming traditional RL approaches that rely on explicit perception preprocessing.

\begin{figure}[t]
    \centering
    \includegraphics[width=0.95\linewidth]{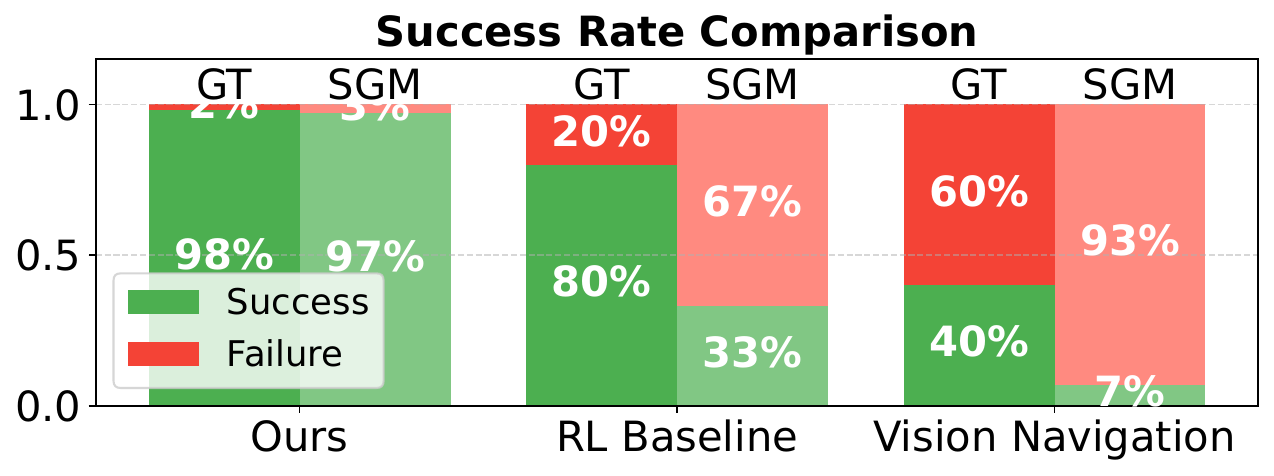}
    \caption{\textbf{Baseline comparison in AirSim.} \rebuttle{We compare our end-to-end vision-based policy with two baselines using two depth inputs: ground-truth (GT) and Semi-Global Matching (SGM). Baselines are: (1) a PPO-based policy with edge-drawing front-end, and (2) a state-of-the-art vision-based navigation method~\cite{Zhang2024BackTN}.}}
    \vspace{-0.8em}
    \label{fig:bl_cmp}
\end{figure}

\rebuttle{
We also compared our approach with state-of-the-art vision-based navigation. Zhang et al.~\cite{Zhang2024BackTN} employed the same differentiable simulation framework and network architecture as ours. However, since their training considers only 3D position, the policy avoids gaps rather than traversing them. When paths were obstructed (Fig.~\ref{fig:simulation}c), this method prone to collisions at high tilt angles, achieving only 40\% success. Using SGM images further increased uncertainty, reducing success to 7\%. In contrast, our method maintained its performance.
}
\subsubsection{Ablation Studies}
After the baseline comparison, we further investigate how different design choices affect performance. 
To this end, we conducted 100 trials for each configuration and reported the resulting position and attitude errors after traversing one or more gaps.
\begin{table}[!hp]
    \centering
    \caption{Average Errors under Different Gap Tilts}
    \label{tab:ablation_combined}
    \begin{tabular}{ccccccc}
        \toprule
        \multirow{2}{*}{\textbf{Gap Tilt}} & \multicolumn{3}{c}{\textbf{Position Error [m]}} & \multicolumn{3}{c}{\textbf{Attitude Error [°]}} \\
        \cmidrule(lr){2-4} \cmidrule(lr){5-7}
        & \textbf{OURS} & \textbf{W/O SG} & \textbf{PM} & \textbf{OURS}& \textbf{W/O SG} & \textbf{PM} \\
        \midrule
        0–30° & \textbf{0.12}  & 0.16 & 0.19 & 3.1 & 4.4 & \textbf{2.8} \\
        30–60° & \textbf{0.15} & 0.46 & 0.16 & \textbf{9.4} & 28.1 & 10.0 \\
        60-80° & \textbf{0.14} & 1.14 & 0.16 & \textbf{11.1} & 55.0 & 26.3 \\
        \bottomrule
    \end{tabular}
\end{table}

Table~\ref{tab:ablation_combined} report position and attitude errors under different gap tilt angles in AirSim, evaluating the contributions of key design choices, including the stop-gradient (SG) operation in (\ref{pos_loss}) and (\ref{rot_loss}), and the CTBR dynamics model versus a point-mass (PM) approximation. The W/O-SG columns show that removing the SG operation significantly increases both position and attitude errors because the policy tends to slow down and hesitate in front of the gap when the target tilt angle is large. Furthermore, we replaced the CTBR model with a PointMass model. Although it achieved accuracy comparable to the CTBR model at smaller gap tilt angles, it exhibited substantially larger rotational errors at higher tilt angles.

\begin{figure}[tp]
    \centering
    \includegraphics[width=\linewidth]{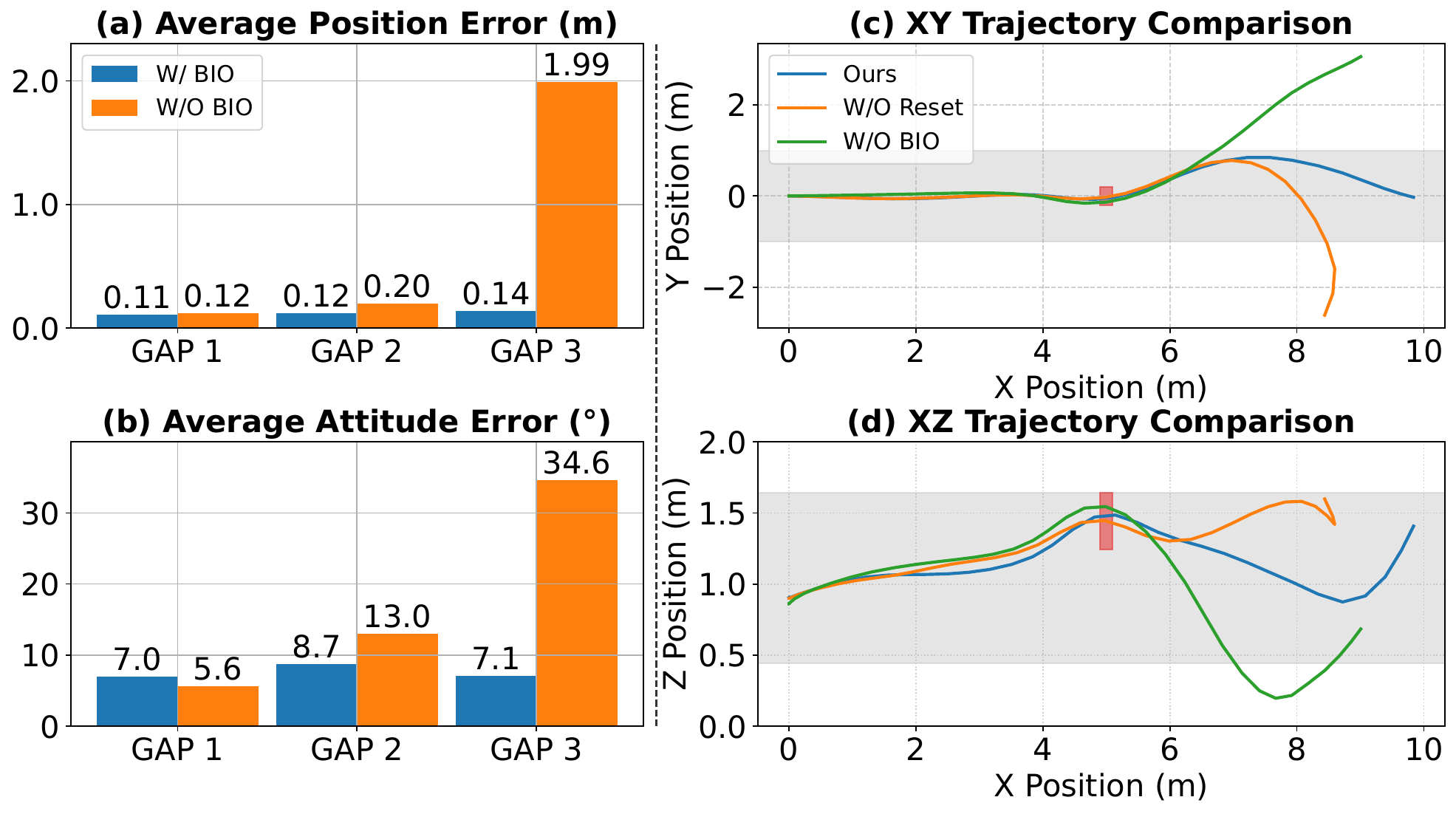}
    \caption{\textbf{Ablation study on the impact of BIO and state reset.} (a-b) Average position and attitude errors across three consecutive gaps. (c-d) Trajectory comparison for validating BIO and gap-crossing prediction.}
    \vspace{-0.8em}
    \label{fig:traj_ablation}
\end{figure}

We further evaluate the effect of the proposed bimodal initialization (BIO) on multi-gap traversal in AirSim. Fig.~\ref{fig:traj_ablation} (a-b) reports position and attitude errors across a three-gap sequence for policies trained with and without BIO. Without BIO, the quadrotor’s post-gap states are highly unstable, leading to rapidly accumulating errors: although first-gap performance is comparable, by the third gap the position and attitude errors reach 1.99 m and 34.59°, often causing crashes. In contrast, policies trained with BIO maintain low errors throughout, demonstrating that bimodal state initialization is crucial for stable multi-gap traversal.

\begin{figure*}[th]
    \centering
    \includegraphics[width=\textwidth]{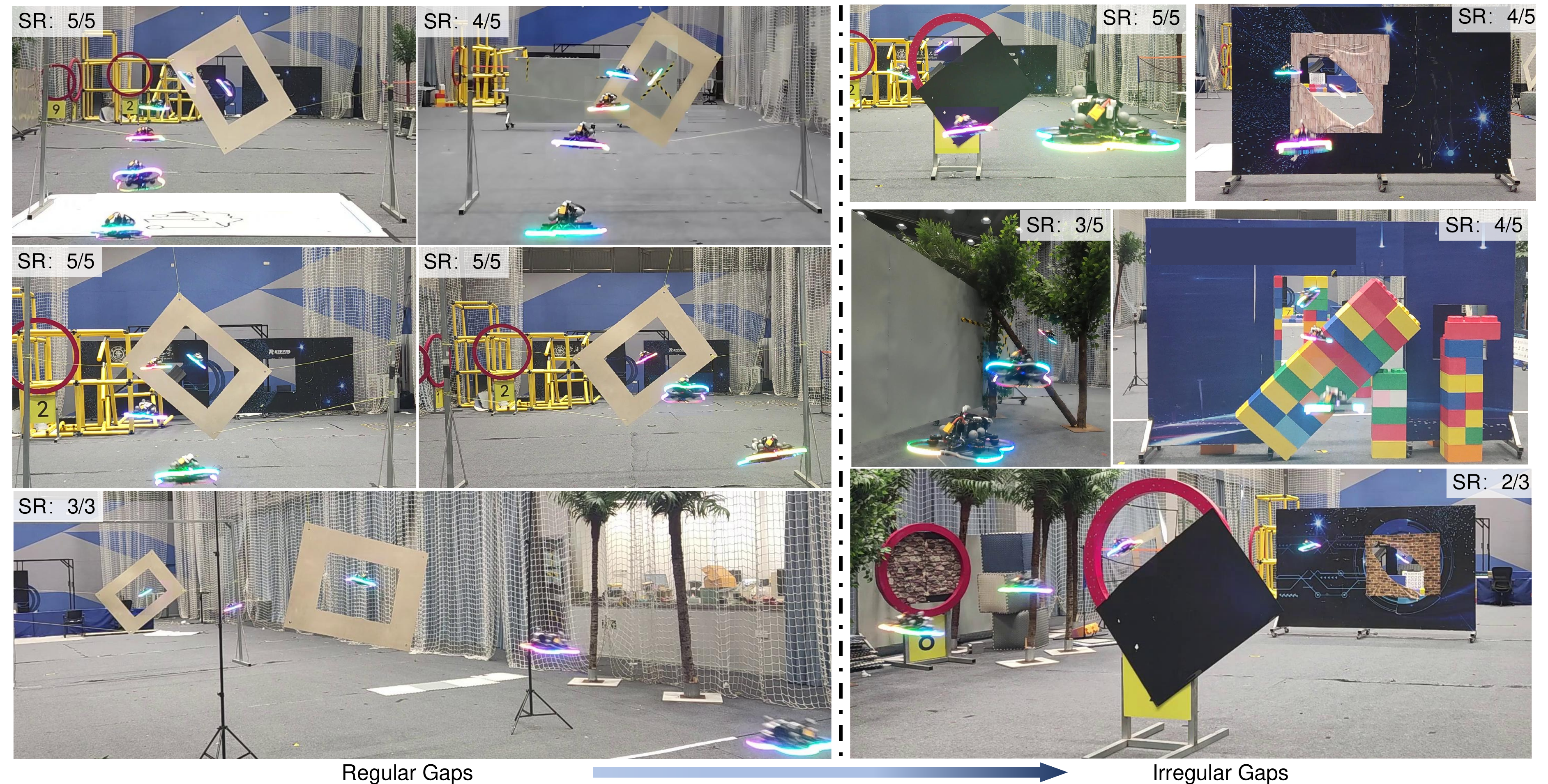}
    \caption{\textbf{Real-world quadrotor experiments in diverse environments.} Top left: regular gaps with varying tilt angles. Top right: irregular gaps, including a half-masked ring, an irregular hole, tiled trees, and a partially occluded door. \rebuttle{The success rate (SR) is annotated on each subplot.}}

    \vspace{-1.0em}
    \label{fig:real_result}
\end{figure*}

To empirically validate that bimodal initialization (BIO) synergizes with the gap-crossing classifier prediction module, we conducted trajectory comparisons in AirSim with three variants: (i) our full method with BIO and hidden-state reset, (ii) without hidden-state reset, and (iii) without BIO. As shown in Fig.~\ref{fig:traj_ablation} (c-d), our method keeps the quadrotor within $y,z \in [-1,1]$ after each gap, whereas the other variants diverge. In particular, without BIO, the quadrotor rapidly loses altitude, reflecting unstable post-gap states. These results confirm that combining hidden-state reset with BIO enables smooth and stable multi-gap traversal without imposing explicit constraints on post-traversal states.

\subsection{Sim-to-Real Transfer}
We deployed the trained policy on a custom-built quadrotor platform (Fig.~\ref{fig:system}, bottom) equipped with an Intel RealSense D435i depth camera and a Radxa Zero3W (1.6GHz quad-core A55 CPU with 1TOPS NPU)\footnote{https://radxa.com/products/zeros/zero3w} for onboard computation. The quadrotor has a diagonal size of 145~mm and a takeoff weight of 462~g. The onboard computer runs Ubuntu~22.04 and ROS2~Humble, with a control frequency set to 30~Hz. The policy network runs on the NPU accelerator of the Radxa Zero3W, achieving an inference time of approximately 5~ms. The quadrotor’s low-level controller operates on a Betaflight flight controller at 1~kHz.

\begin{figure}[ht]
    \centering
    \includegraphics[width=\linewidth]{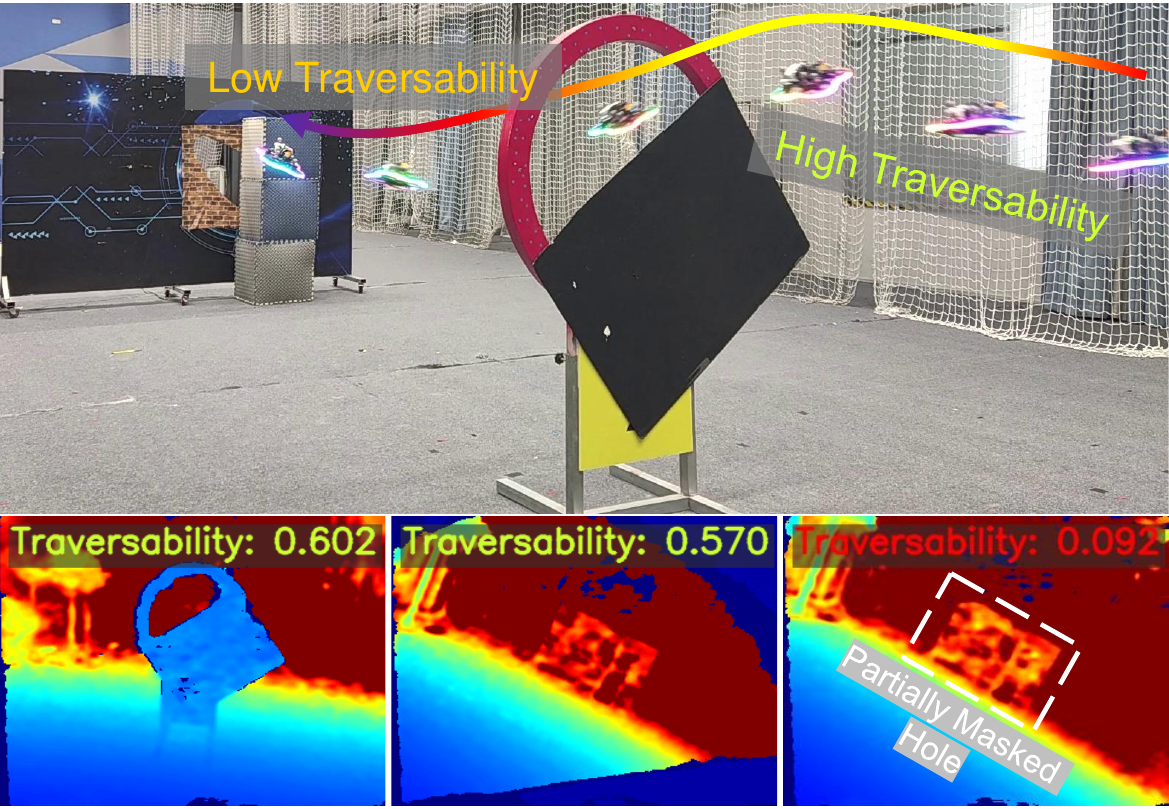}
    \caption{\textbf{Evaluation of the traversability prediction module on the real quadrotor.} The quadrotor first successfully traverses the gap, while the predicted traversability score remains high. When the target gap is switched to an irregular hole partially occluded by an obstacle, the predicted score drops sharply, triggering an emergency stop.}
    \vspace{-0.8em}
    \label{fig:collide_risk}
\end{figure}

We tested the quadrotor in diverse real-world environments, including a training-like scenario with regular gap and varying tilt angles (Fig.~\ref{fig:real_result}, left), and generalization scenarios with irregular gaps, such as a half-masked ring, an irregular hole, tiled trees, and a door partially blocked by a tiled wall (Fig.~\ref{fig:real_result}, right). Consecutive gap traversal was also evaluated in both training-like and generalization settings (Fig.~\ref{fig:real_result}, bottom), with a desired velocity of 2.5~m/s. Notably, the policy was trained purely in simulation using regular gap shapes, yet it successfully generalized to significantly different real-world scenarios. These results demonstrate the robustness of the learned end-to-end policy and indicate that the network has internalized generalizable strategies for agile gap traversal beyond the training distribution. \rebuttle{We evaluate performance across diverse environments using the success rate (SR) and find that our policy maintains high success rates even in scenarios with irregular gaps.}

We also evaluated the traversability prediction module on the real quadrotor to validate its performance in real-world conditions. As shown in Fig.~\ref{fig:collide_risk}, the quadrotor first successfully traverses the half-masked ring, with the predicted traversability score remaining high. When the target gap is switched to an irregular hole partially occluded by an obstacle, the predicted traversability drops sharply, which triggers the state machine to execute an emergency stop.

These real-world experiments confirm that our framework not only transfers seamlessly from simulation to hardware but also generalizes effectively to unseen scenarios, thereby validating its practical applicability.

\subsection{Additional Experiments}
Finally, to further validate the robustness and reliability of our method, we conducted supplementary experiments. These studies examine two key aspects: (i) the policy’s ability to handle uncertainty in the target direction, and (ii) the accuracy of the traversability predictor under varying model capacities, thereby complementing the preceding analyses.

\begin{figure}[t]
    \centering
    \includegraphics[width=\linewidth]{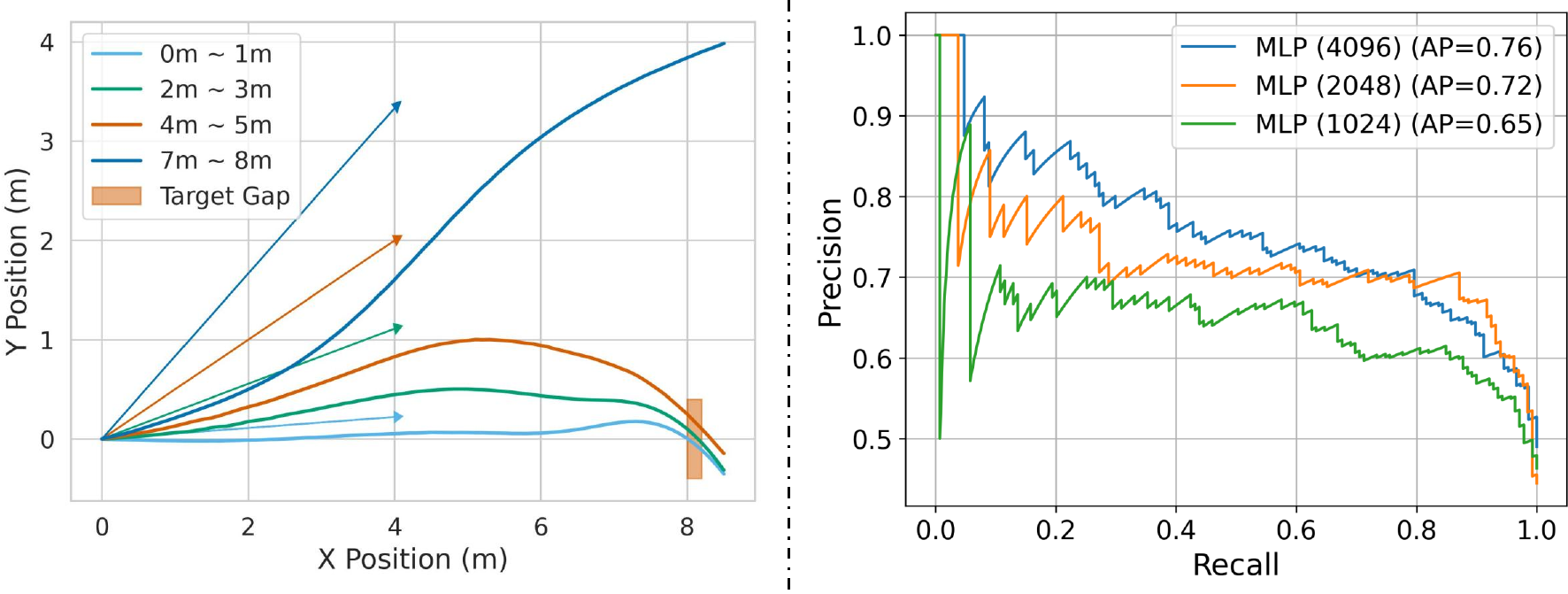}
    \caption{\textbf{Additional experiments.} Left: average trajectories under different target direction noise levels. Right: PR curves of the collision traversability predictor under different model sizes.}
    \label{fig:supplementary}
    \vspace{-0.8em}
\end{figure}

\paragraph*{\textbf{Effect of Target Direction Uncertainty}}
We evaluated the policy’s robustness to imprecise target velocity inputs by perturbing the target direction $ \Tilde{v}_\text{target}$ with varying noise levels in simulation. With small perturbations, the quadrotor consistently flew straight through the gap. Under moderate noise (2–3 m), it still reliably traversed the gap. For larger noise (4–5 m), the quadrotor initially followed the noisy direction but reoriented toward the actual gap once it entered the field of view. Only extremely large perturbations (7–8 m) caused traversal failure. These results indicate that $ \mathbf{\Tilde{v}}_\text{target}$ functions primarily as a coarse directional cue and does not need precise specification during deployment.

\paragraph*{\textbf{Traversability Prediction Accuracy}}
We evaluated the predictor with hidden-layer sizes of 1024, 2048, and 4096. A dataset of 300 AirSim trajectories was collected, where gap sizes were randomly scaled to 0.5–1, ensuring that collision-free trajectories accounted for half of the total samples. Precision–recall (PR) curves and average precision (AP) scores were computed based on predicted traversability before gap traversal. Larger networks achieved higher AP (0.64, 0.72, 0.76). However, the performance gain from MLP(2048) to MLP(4096) was marginal, indicating that the predictor’s performance tends to saturate as the model size increases.

\section{Conclusion}
This work presents a fully vision-based, end-to-end framework for agile quadrotor navigation through narrow and irregular gaps. By leveraging differentiable simulation, Stop-Gradient optimization, and bimodal state initialization, the policy directly maps depth images to control commands, enabling stable multi-gap traversal. Auxiliary prediction modules for gap-crossing and traversability further improve reliability and anticipatory hazard avoidance. Extensive simulation and real-world experiments demonstrate robust sim-to-real transfer, generalization to unseen environments, and practical applicability for agile UAV navigation.

Future work includes integrating obstacle avoidance with the gap-traversal policy for simultaneous navigation and investigating fully self-adaptive traversal using only an obstacle-avoidance loss.

\ifCLASSOPTIONcaptionsoff
  \newpage
\fi

\bibliographystyle{ieeetr}
\bibliography{reference}

\end{document}